\documentclass[letterpaper, 10 pt, journal, twoside]{ieeetran}
\IEEEoverridecommandlockouts    % This command is only needed if
\usepackage{hyperref}
\usepackage{makecell}
\usepackage{threeparttable}
\usepackage{caption}
\usepackage{soul}
\usepackage{color}
\usepackage{array}
\usepackage[table]{xcolor}
\usepackage{colortbl}
\usepackage{stfloats}
\usepackage{pifont} 
\hypersetup{
    colorlinks=true,
    linkcolor=magenta,
    filecolor=magenta,      
    urlcolor=magenta,
}

\usepackage{graphics}           
\usepackage{times}              
\usepackage{amsmath}            
\usepackage{amssymb}            
\usepackage{graphicx}
\usepackage{algorithm}
\usepackage[noend]{algpseudocode}
\usepackage{booktabs}
\usepackage{color}
\definecolor{instructioncolor}{rgb}{.5,.5,.5}

\usepackage[font=small]{caption}

\def\secref#1{Sec.~\ref{#1}}
\def\figref#1{Fig.~\ref{#1}}
\def\tabref#1{Tab.~\ref{#1}}
\def\eqref#1{Eq.~(\ref{#1})}

\makeatletter
\usepackage{xspace}
\DeclareRobustCommand\onedot{\futurelet\@let@token\@onedot}
\def\@onedot{\ifx\@let@token.\else.\null\fi\xspace}

\makeatother

\usepackage{array}
\newcolumntype{L}[1]{>{\raggedright\let\newline\\\arraybackslash\hspace{0pt}}m{#1}}
\newcolumntype{C}[1]{>{\centering\let\newline\\\arraybackslash\hspace{0pt}}m{#1}}
\newcolumntype{R}[1]{>{\raggedleft\let\newline\\\arraybackslash\hspace{0pt}}m{#1}}

\usepackage{bm}
\usepackage{multirow}
\usepackage{rotating}  % for rotating sideway

\title{RLPR: Radar-to-LiDAR Place Recognition via Two-Stage Asymmetric Cross-Modal Alignment for Autonomous Driving}

\author{Zhangshuo Qi$^{1}$, Jingyi Xu$^{2}$, Luqi Cheng$^{1}$, Shichen Wen$^{1}$, and Guangming Xiong$^{1*}$%
\thanks{Manuscript received: March, 8, 2026; Revised April, 23, 2026; Accepted June, 9, 2026.}%Use only for final RAL version
\thanks{This paper was recommended for publication by Editor Ayoung Kim upon evaluation of the Associate Editor and Reviewers’ comments.
This work was supported by the National Natural Science Foundation of China under Grant 52372404.} %Use only for final RAL version
\thanks{$^{1}$Zhangshuo Qi, Luqi Cheng, Shichen Wen and Guangming Xiong are with Beijing Institute of Technology, Beijing, 100081, China.}%
\thanks{$^{2}$Jingyi Xu is with Shanghai Jiaotong University, Shanghai, 200240, China.}%
\thanks{$^{*}$Corresponding author: Guangming Xiong (xiongguangming@bit.edu.cn)}%
\thanks{Digital Object Identifier (DOI): see top of this page.}
}

\begin{document}
\maketitle

\IEEEpeerreviewmaketitle
% \thispagestyle{empty}
% \pagestyle{empty}

%%%%%%%%%%%%%%%%%%%%%%%%%%%%%%%%%%%%%%%%%%%%%%%%%%%%%%%%%%%%%%%%%%%%%%%%%%%%%%%%
\begin{abstract}

All-weather autonomy is critical for autonomous driving, which necessitates reliable localization across diverse scenarios. While LiDAR place recognition is widely deployed for this task, its performance degrades in adverse weather. Radar is inherently weather-resilient; however, in scenarios where LiDAR-based maps already serve as the foundational infrastructure, constructing radar maps at scale would incur non-trivial additional overhead. To leverage the existing infrastructure with weather-robust onboard sensors, radar-to-LiDAR place recognition has garnered increasing interest. However, extracting discriminative and generalizable features shared between modalities remains challenging. In this work, we propose RLPR, a robust radar-to-LiDAR place recognition framework compatible with both scanning radars and phased-array radars. We design a dual-stream network that matches solely on shared spatial structures, deliberately discarding modality-specific signatures such as RCS and Doppler velocity in exchange for compatibility across radar types. Subsequently, motivated by our task-specific asymmetry observation between radar and LiDAR, we introduce a two-stage asymmetric cross-modal alignment (TACMA) strategy, which leverages the pre-trained radar branch as a discriminative anchor to guide the alignment process. Experiments on five datasets demonstrate that RLPR achieves state-of-the-art recognition accuracy with strong zero-shot generalization capabilities. Our open-source code is available at \url{https://github.com/QiZS-BIT/RLPR}.
\end{abstract}

%%%%%%%%%%%%%%%%%%%%%%%%%%%%%%%%%%%%%%%%%%%%%%%%%%%%%%%%%%%%%%%%%%%%%%%%%%%%%%%%
\section{Introduction}
\label{sec:intro}

\IEEEPARstart{I}{n} autonomous driving, all-weather localization is critical for operational safety. Currently, LiDAR place recognition (LPR) has matured into a de facto standard for global localization in GPS-denied environments~\cite{uy2018pointnetvlad, ma2022overlaptransformer, luo2024bevplace++}, typically by matching query scans against a map database. However, the performance of LPR degrades substantially in adverse weather~\cite{qi2025lrfusionpr, luang2025reslpr}, where radar place recognition (RPR) excels due to its inherent weather resilience~\cite{cait2022autoplace, gadd2024open, peng2024transloc4d}. Nonetheless, as many autonomous driving scenarios already utilize high-fidelity LiDAR maps as foundational infrastructure, constructing dedicated radar maps at scale specifically for localization under adverse weather would incur a non-trivial overhead. Radar-to-LiDAR (R2L) place recognition thus emerges as a practical solution for scenarios with vast available LiDAR infrastructure but lacking radar maps, enabling weather-robust localization~\cite{yin2021radar, nayak2024ralf}.

Despite its potential, the practical deployment of R2L methods is hindered by the substantial cross-modal gap between these two modalities. Specifically, several critical challenges remain to be solved: (1) Existing R2L methods are primarily tailored for scanning radar~\cite{yin2021radar, nayak2024ralf}. However, as phased-array radars become increasingly prevalent in autonomous driving, spanning 3D (exemplified by the Continental ARS-408) and 4D variants, an R2L framework capable of accommodating different radar types remains underexplored. (2) Current R2L approaches typically adopt a symmetric alignment paradigm, where both modalities are treated with equal priority, forcing them to converge toward a mutual optimization target. However, this paradigm often overlooks the preservation of intra-modal discriminability while simultaneously underestimating the heterogeneity of different modalities, which can pose challenges for learning sufficiently generalizable representations.

\begin{figure}
%\vspace{0.2cm}
  \centering
  \includegraphics[width=1\linewidth]{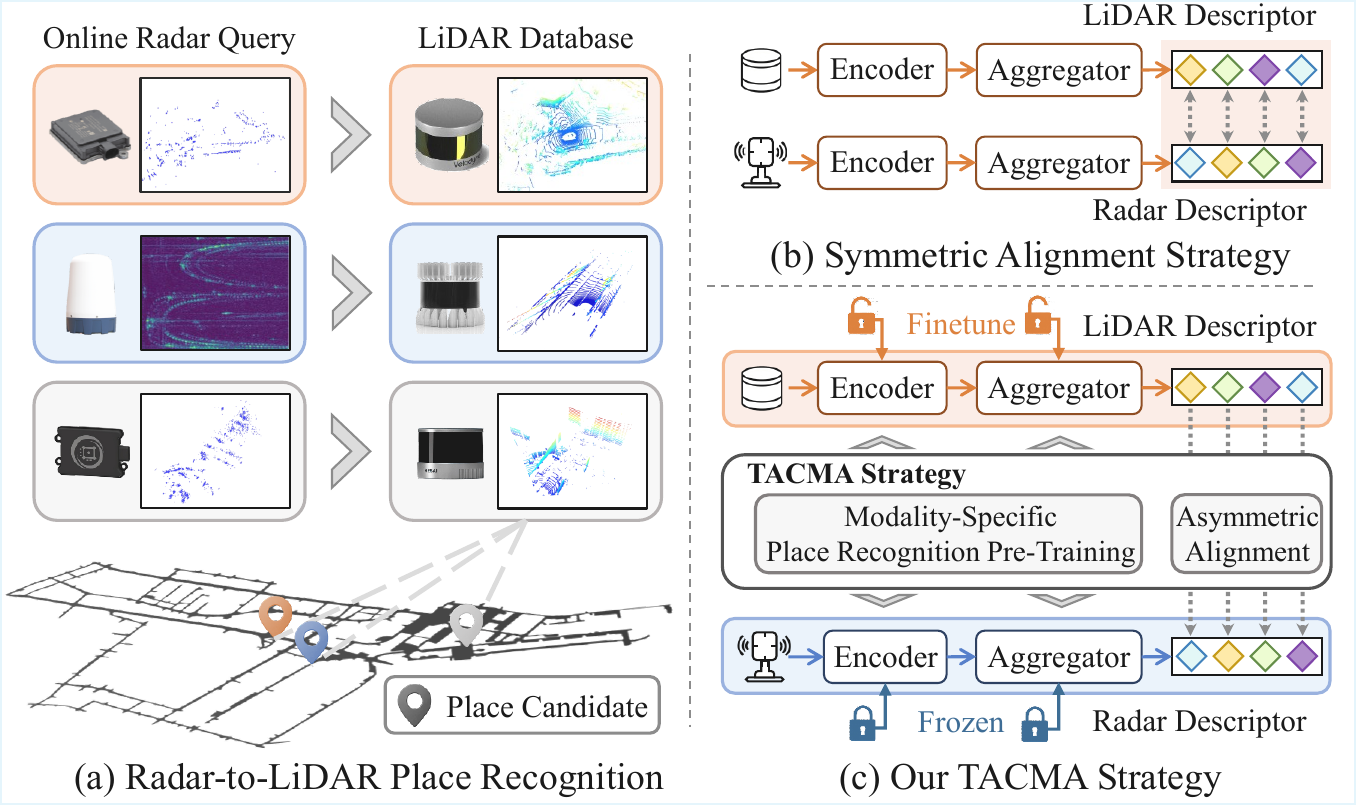}
  \caption{Conceptual overview of the proposed RLPR framework. (a) RLPR facilitates robust place recognition across radar types. (b) Existing methods perform symmetric feature alignment. (c) Motivated by the task‑specific asymmetry between radar and LiDAR, we design a two-stage asymmetric cross-modal alignment (TACMA) strategy that enables effective feature alignment.}
  \label{fig:motivation}
  \vspace{-0.75cm}
\end{figure}

To address these challenges, we propose RLPR, a radar-to-LiDAR place recognition framework capable of accommodating both scanning radars and phased-array variants. Our framework prioritizes polar BEV as a generalized representation over modality-specific physical signatures, consciously trading sensor type-specific depth for a compatibility grounded in shared spatial structures (see~\secref{sec:network_architecture}). We then design a network architecture to generate dual-descriptors, extracting both local and global geometric features for cross-modal alignment. Furthermore, we re-examine the symmetric alignment paradigm through the lens of entropy evolution in LiDAR and radar features during place recognition pre-training, identifying an inherent task-specific asymmetry between LiDAR and radar. Guided by this diagnostic heuristic, we propose a two-stage asymmetric cross-modal alignment (TACMA) strategy, as shown in~\figref{fig:motivation}. Following independent pre-training, we freeze the radar branch as an anchor, streamlining feature alignment along a more tractable path while anchoring the shared latent space onto a discriminative manifold, facilitating effective cross-modal retrieval.

In summary, our main contributions are as follows:
\begin{itemize}
\item We propose RLPR, a R2L place recognition framework that leverages shared spatial structures to achieve compatibility across common radar types in autonomous driving (see~\secref{sec:network_architecture}), ranging from scanning radars to 3D and 4D phased-array variants.
\item We conduct a systematic investigation into R2L cross-modal alignment, leading to a principled two-stage asymmetric alignment strategy to bridge the sensor-specific modality gap.
\item Extensive evaluations across five datasets demonstrate that RLPR achieves accurate and generalizable R2L place recognition, while exhibiting significant robustness in adverse weather conditions.
\end{itemize}

%%%%%%%%%%%%%%%%%%%%%%%%%%%%%%%%%%%%%%%%%%%%%%%%%%%%%%%%%%%%%%%%%%%%%%%%%%%%%%%%
\section{Related Work}
\label{sec:related}

\subsection{LiDAR and Radar Place Recognition}
\label{sec:lrpr}
The widespread adoption of LiDAR maps in autonomous driving has catalyzed significant progress in LiDAR place recognition (LPR)~\cite{ma2022overlaptransformer, luo2024bevplace++, komorowski2022improving, ma2023cvtnet}. We recommend referring to~\cite{yin2025general} for a comprehensive survey. Despite their precision, LiDARs are prone to noise in adverse weather, leading to performance degradation in LPR systems~\cite{qi2025lrfusionpr, luang2025reslpr}. 

In contrast, the inherent robustness of radar has sparked increasing interest in radar place recognition (RPR). Many existing methods have primarily leveraged scanning radars for their relatively dense and accurate nature. For example, KidnappedRadar~\cite{suaftescu2020kidnapped} learns to extract rotation-invariant features, ReFeree~\cite{kim2024referee} utilizes statistical methods for lightweight descriptor generation, RaPlace~\cite{jang2023raplace} employs the Radon transform for robust representations, and Open-RadVLAD~\cite{gadd2024open} operates at the radial frequencies for feature clustering. However, scanning radars are often constrained by their hardware cost and physical footprint. Phased-array radars provide a more deployable alternative, where the 3D and 4D subcategories, classified by output dimensionality, have both seen extensive application in autonomous driving. For instance, AutoPlace~\cite{cait2022autoplace} achieves place recognition on the low-resolution and elevation-limited ARS-408 3D radar via temporal feature accumulation, and TransLoc4D~\cite{peng2024transloc4d} explores the effectiveness of sparse convolution in processing 4D radar point clouds. Recently, to bridge the gap between disparate radar types, SHeRLoc~\cite{kim2025sherloc} extends the paradigm to heterogeneous radar place recognition.

These advances establish RPR as the logical choice for all-weather localization when radar maps are available. However, where extensive LiDAR infrastructure already exists, constructing dedicated radar maps at scale involves substantial overhead. This positions R2L place recognition as a complementary solution for scenarios where existing LiDAR maps are queried by onboard radar sensors for reasons of cost or weather robustness.

\subsection{Cross-Modal Place Recognition}
\label{sec:cpr}
Cross-modal place recognition primarily evolved from camera-to-LiDAR (C2L) matching. Cattaneo et al.~\cite{cattaneo2020global} laid the foundation for C2L by aligning 2D image and 3D point cloud features in a shared latent space. To mitigate the modality gap, subsequent studies have explored projecting point clouds into range images. For instance, LIP-Loc~\cite{puligilla2024liploc} adopts a contrastive learning paradigm, and I2P-CMPR~\cite{yao2025i2pcmpr} further captures 2D cross-modal correspondences at both global and local scales.

Drawing inspiration from C2L, Radar-to-LiDAR~\cite{yin2021radar} pioneered radar-to-LiDAR (R2L) place recognition by exploring cross-modal matching between scanning radar and LiDAR BEV, while RaLF~\cite{nayak2024ralf} attempts to improve generalization by scaling up the training corpus. Moving beyond existing R2L methods tailored for scanning radars, in this work, we extend R2L capability to accommodate different radar types by matching on shared spatial structures (see \secref{sec:network_architecture}). Furthermore, we introduce an asymmetric alignment strategy for R2L cross-modal alignment. Our asymmetric alignment strategy is related to~\cite{cattaneo2020global}, which first showed that freezing one modality can be effective for C2L place recognition. However, we revisit it as a task-specific alignment problem in R2L and systematically study how pre-training, alignment direction, and loss objective affect cross-modal retrieval.

Regarding the architecture, we draw inspiration from the efficient feature extraction pipeline of LRFusionPR~\cite{qi2025lrfusionpr}, adopting its polar BEV representation and ResNet backbone. However, while LRFusionPR targets multimodal fusion, our work addresses the distinct challenge of cross-modal R2L retrieval. Beyond the shared encoding modules, our primary contribution lies in the architecture specifically designed for the R2L task, along with the cross-modal alignment strategy.

%%%%%%%%%%%%%%%%%%%%%%%%%%%%%%%%%%%%%%%%%%%%%%%%%%%%%%%%%%%%%%%%%%%%%%%%%%%%%%%%
\section{Our Approach}
\label{sec:approach}

\subsection{System Overview}
\label{sec:overview}
% Given a query radar frame and a database consisting of LiDAR scans, our objective is to retrieve the most similar LiDAR scan corresponding to the same place. The core challenge of R2L place recognition lies in identifying cross-modal commonalities to bridge the domain gap. To this end, we propose RLPR, the overall pipeline of which is illustrated in~\figref{fig:overall}. In light of the inherent asymmetry and discriminability of different modalities, we design a dual-stream architecture with identical configurations but independent parameter spaces. Data from different modalities are first projected into polar BEV representations. Before entering the backbone, these representations are processed by the proposed Polar Context Enhancer (PCE) module to filter noise and bridge modality discrepancies. The backbone then extracts geometric features from the refined representations, which are subsequently aggregated into dual-descriptors to capture both global context and local details. Finally, the two branches of the network are aligned through our proposed two-stage asymmetric cross-modal alignment (TACMA) strategy. We first leverage place recognition pre-training to establish intra-modal discriminability, followed by an asymmetric alignment stage where the radar branch serves as an anchor to uncover generalizable commonalities.

\begin{figure*}[ht]
%\vspace{0.2cm}
  \centering
  \includegraphics[width=0.95\linewidth]{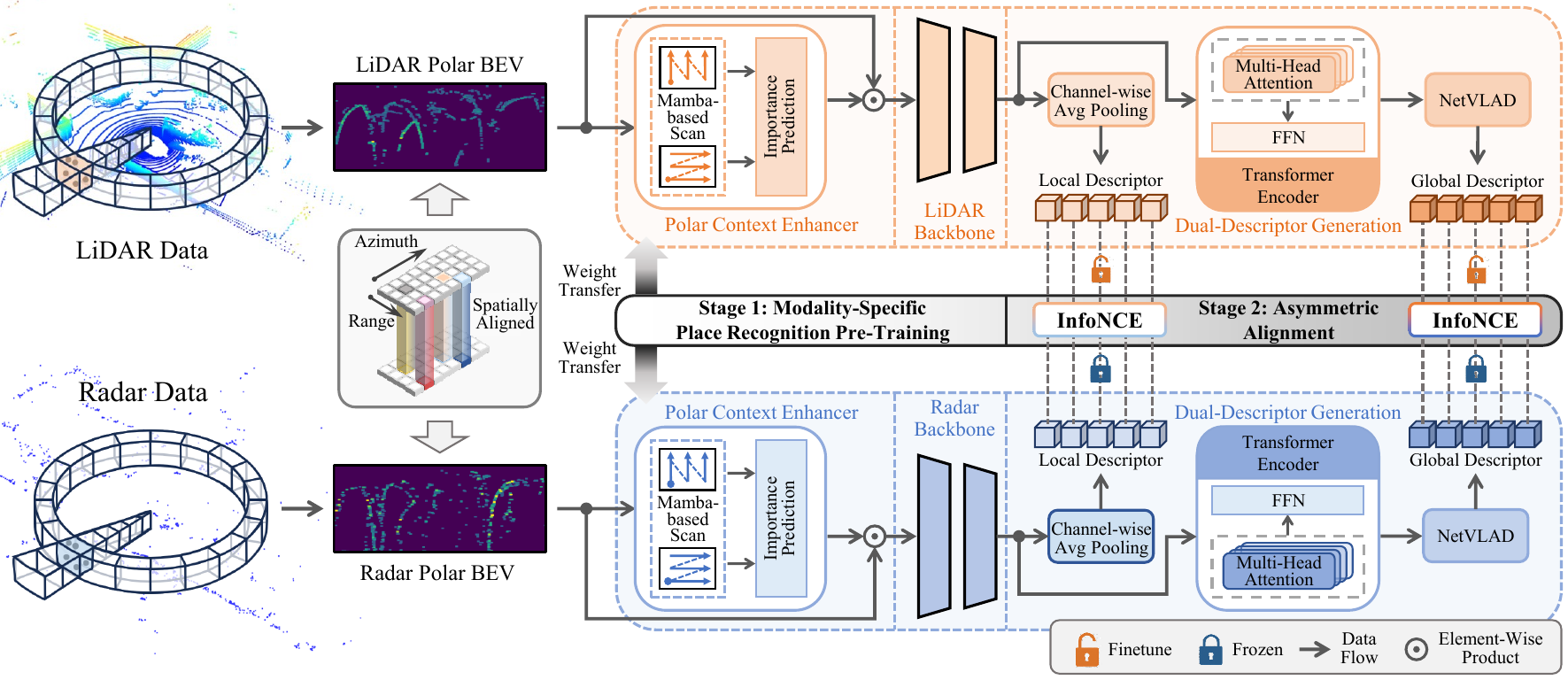}
  \caption{The system overview of RLPR. The proposed radar-to-LiDAR place recognition framework consists of a dual-stream network and the two-stage asymmetric cross-modal alignment (TACMA) strategy. The dual-stream network is designed to extract global and local geometric features, while the TACMA strategy exploits cross-modal commonalities for accurate and generalizable retrieval.}
  \label{fig:overall}
  \vspace{-0.7cm}
\end{figure*}

Given a radar query and a LiDAR database, our objective is to retrieve the LiDAR scan corresponding to the same place. The overall pipeline of our proposed RLPR is illustrated in~\figref{fig:overall}. First, raw data are projected into polar BEV representations and refined by the Polar Context Enhancer (PCE) to mitigate noise. The backbone then extracts geometric features, which are aggregated into dual-descriptors capturing both global and local context. Specifically, we adopt a homogeneous two-branch design to provide a structural guarantee for embedding space similarity, while serving as a control to isolate inherent modality differences from architectural biases. Finally, we employ a Two-stage Asymmetric Cross-modal Alignment (TACMA) strategy. This approach first establishes intra-modal discriminability via modality-specific pre-training, followed by an asymmetric alignment where the radar branch serves as an anchor to uncover generalizable commonalities.

\subsection{Network Architecture}
\label{sec:network_architecture}
\textbf{Polar BEV Representation:} RLPR exploits the widely-recognized polar BEV representation~\cite{qi2025lrfusionpr, gadd2024open, yin2021radar} as a viewpoint-robust data foundation for cross-modal place recognition. Given a point cloud $\mathcal{P}$ from either LiDAR or radar, we discretize the 3D space into a 2D polar grid $\mathcal{B}$. Specifically, each point $p_{i}=\left(x_{i}, y_{i}, z_{i}\right)^{\text{T}}$ in the point cloud $\mathcal{P}$ is mapped to the pixel coordinates $\left(u_{i}, v_{i}\right)$ of $\mathcal{B}$ by:
\begin{align}
    \left( \begin{array}{c} 
    u_{i} \\ 
    v_{i} 
    \end{array} \right) 
    &= \left( \begin{array}{c} 
    \frac{1}{2} \left[ 1 - \pi^{-1} \arctan\left(y_{i}, x_{i} \right) \right] w \\ 
    \left[ m^{-1} \sqrt{x_{i}^2 + y_{i}^2} \right] h
    \end{array} \right) \label{eq:lidarp}
\end{align}
where $w$ and $h$ represent the width and height of the polar BEV $\mathcal{B}$, and $m$ is a predefined maximum perception range. The value at position $\left(u_{i}, v_{i}\right)$ corresponds to the number of points falling within the corresponding grid cell. Notably, an explicit trade-off exists in this design, as the polar BEV discards modality-specific physical signatures, such as the Doppler velocity that most distinguishes 4D radar. Consequently, while this representation remains applicable to 4D radar, there is still substantial room for improvement in its specialized performance. Despite these omissions, the polar BEV effectively preserves the spatial structures for R2L matching, providing a viable foundation for learning cross-modal geometric correlations in a generalizable manner.
%Notably, the polar representation naturally accommodates the horizontal field-of-view (FoV) variations. For radars with limited FoV, we adjust the mapping parameters accordingly to concentrate on an azimuthal sub-region, thereby preserving the high-fidelity structural details. By mapping raw data into this unified format, we effectively abstract away modality-specific physical signatures, preserving only the geometric structure. This establishes a common ground that enhances compatibility across radar types. We acknowledge that this unification is a deliberate trade-off: by collapsing Doppler, RCS, and elevation into a count map, the representation acts as a principled common denominator for alignment with LiDAR's geometric space rather than exploiting modality-specific cues that are not the operative information for R2L.
%Moreover, while existing works~\cite{qi2025lrfusionpr, yin2021radar} primarily utilized polar BEV for sensors with a panoramic FoV, we demonstrate that it is also suited for sensors with a constrained FoV, such as 4D radars.

\textbf{Polar Context Enhancer:} To mitigate inherent radar noise, RLPR introduces a lightweight Mamba-based enhancer that gates the polar BEV input.
% To facilitate the processing of these sequences, the SSM is formulated as the following discrete-time recurrence:
% \begin{align}
%     h_k &= \mathbf{\bar{A}}h_{k-1} + \mathbf{\bar{B}}x_k \\
%     y_k &= \mathbf{C}h_k + \mathbf{D}x_{k}
% \end{align}
% where $\mathbf{\bar{A}}$, $\mathbf{\bar{B}}$, $\mathbf{C}$, and $\mathbf{D}$ are the discretized system parameters, as detailed in~\cite{gu2024mamba}. 
While standard Mamba layer~\cite{gu2024mamba} is designed for modeling 1D sequences, the polar BEV $\mathcal{B}$ exhibits a structured geometric topology along the range and azimuth axes. To capture global dependencies across these axes, we introduce a two-way scanning strategy. Specifically, after patchifying the polar BEV $\mathcal{B}$ as $\mathcal{B}_{\text{patch}} \in \mathbb{R}^{H\times W \times C}$, we construct two mutually orthogonal input sequences $s_{\text{rng}} \in \mathbb{R}^{(H \cdot W) \times C}$ and $s_{\text{azi}} \in \mathbb{R}^{(W \cdot H) \times C}$ to traverse the grid.
% \begin{align}
%     x_{\text{rng}} &= \text{Flatten}(\mathcal{B_{\text{patch}}}) \in \mathbb{R}^{(H \cdot W) \times C} \\
%     x_{\text{azi}} &= \text{Flatten}(\mathcal{B}_{\text{patch}}^{\text{T}}) \in \mathbb{R}^{(W \cdot H) \times C}
% \end{align}
% where $\text{Flatten}(\cdot)$ denotes the vectorization operation. 
%The sequences are then processed by the Mamba layer to aggregate dependencies along both axes. 
For efficiency, we propose using the non-causal bidirectional Mamba from~\cite{hatamizadeh2025mambavision}, precluding the need for separate backward scans.

\begin{figure}[t]
%\vspace{0.2cm}
  \centering
  \includegraphics[width=1\linewidth]{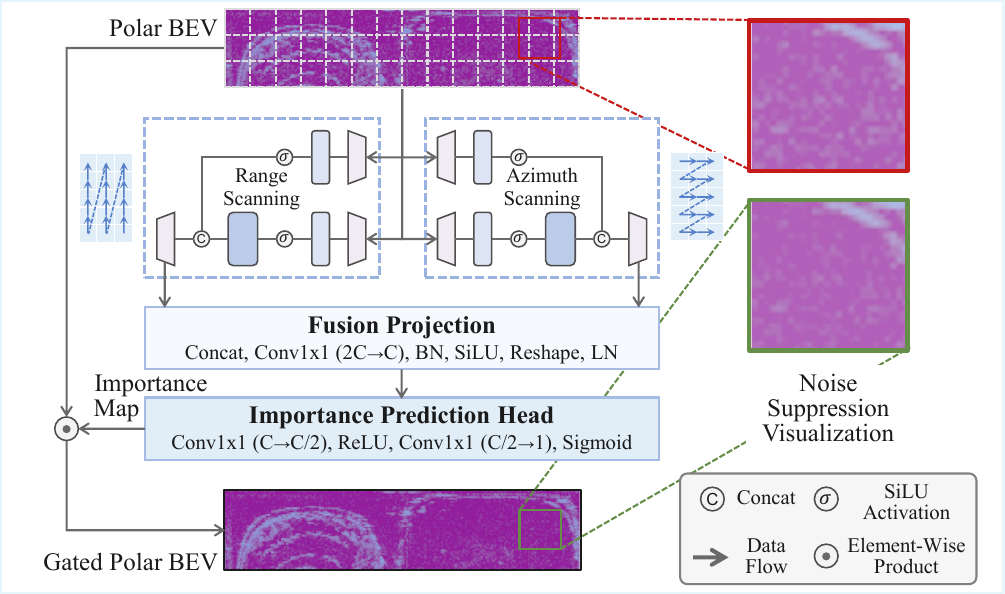}
  \caption{Architecture of the proposed Polar Context Enhancer. It captures global dependencies through two-way scanning along range and azimuth axes, generating an importance map to gate the polar BEV. The corresponding images share the same color scale.}
  \label{fig:mamba}
  \vspace{-0.75cm}
\end{figure}

Finally, the features are concatenated and passed through 2D convolutions followed by a sigmoid activation to produce the importance map, which suppresses noise-prone regions via element-wise gating, as illustrated in~\figref{fig:mamba}. Furthermore, for the LiDAR branch, the PCE module ensures latent‑space consistency across modalities, and serves to attenuate the domain gap in the alignment process.

\textbf{Dual-Descriptor Generation:} After processing via the PCE module and a modality-specific backbone consisting of two ResBlocks~\cite{he2016deep}, the polar BEV $\mathcal{B}$ is compressed into a feature map $\mathcal{F} \in \mathbb{R}^{H' \times W' \times C'}$. The dual‑descriptor generation process then aggregates the geometric information at both local and global levels. 
%As detailed in~\secref{sec:ablation}, cross-modal alignment benefits significantly from combining local geometric consistency and holistic scene context.

To preserve geometric details while ensuring cross-modal alignment stability, we apply channel-wise average pooling (CAP) to the feature map $\mathcal{F}$, computing $\frac{1}{C'}\sum_{c=1}^{C'}\mathcal{F}_{u,v,c}$ at each spatial position $(u, v)$. A linear projection then yields the local descriptor $\mathcal{D}_{\text{loc}}\in\mathbb{R}^{D}$. Unlike global pooling, which reduces along the spatial axes, CAP retains the spatial topology and structural cues in the polar BEV feature map. Among channel-wise alternatives, average pooling is preferred over channel-wise max pooling (CMP). This is because the per-position maximum is dominated by the single strongest channel activation, which tends to be inconsistent across radar and LiDAR features; averaging instead yields a channel-consensus response that provides more stable supervision for the cross-modal objective (see~\tabref{tab:architecture}).

To capture the holistic context, we employ a transformer encoder~\cite{vaswani2017attention} to process the feature map. The self-attention mechanism can be formulated as:
\begin{align}
    \mathrm{Attention}(\mathcal{Q},\mathcal{K},\mathcal{V})=\mathrm{softmax}\left(\frac{\mathcal{Q}\mathcal{K}^{\mathrm{T}}}{\sqrt{d_{\text{k}}}}\right)\mathcal{V}
\end{align}
where $\mathcal{Q},\mathcal{K},\mathcal{V}$ are linear projections of the feature map $\mathcal{F}$, representing queries, keys, and values, respectively, and $d_{\text{k}}$ is the channel dimension. Following the transformer encoder, a NetVLAD~\cite{arandjelovic2016netvlad} layer is employed to aggregate the context into the global descriptor $\mathcal{D}_{\text{glob}} \in \mathbb{R}^{D}$. Finally, $\mathcal{D}_{\text{loc}}$ and $\mathcal{D}_{\text{glob}}$ are concatenated as $\mathcal{D} \in \mathbb{R}^{2D}$ for cross-modal retrieval.

\subsection{Two-Stage Cross-Modal Alignment}
\label{sec:asymmetric_cross_modal_alignment}
\textbf{Modality-Specific Place Recognition Pre-Training:} Given the proposed network architecture, the efficacy of R2L retrieval hinges on the seamless alignment of disparate feature manifolds. However, we observe that directly enforcing cross-modal alignment, employing simultaneous joint optimization, or pre-training only a single modality as a fixed teacher~\cite{cattaneo2020global} often leads to suboptimal results (see~\tabref{tab:pretraining}).

To address this, we decouple the learning process into a two-stage strategy. In the first stage, both branches are independently pre-trained to extract stable and discriminative features. Following the established practice in LPR~\cite{luo2024bevplace++, qi2025lrfusionpr, komorowski2022improving}, we employ the lazy triplet loss~\cite{uy2018pointnetvlad} for place recognition pre-training. Taking the radar branch as an example, for a given query descriptor $\mathcal{D}^{\text{ra}}_{\text{q}}$, we randomly sample a positive sample $\mathcal{D}^{\text{ra}}_{\text{pos}}$ and select $J$ hard negatives $\mathcal{D}_{\text{neg}}^{\text{ra}}$ via hard negative mining to form a triplet. The lazy triplet loss $\mathcal{L}_{\text{LT}}$ is then formulated as:
\begin{align}
    \mathcal{L}_{\text{LT}}\!=\!\Big[\alpha\!+\!d(\mathcal{D}^{\text{ra}}_{\text{q}},\mathcal{D}^{\text{ra}}_{\text{pos}})\!-\!\mathop{\max}_{j \in \{1,\dots,J\}} d(\mathcal{D}^{\text{ra}}_{\text{q}},\mathcal{D}_{\text{neg},j}^{\text{ra}})\Big]_+
\end{align}
where $\left [ \cdots \right ]_+$ denotes the hinge loss, $d(\cdot)$ means the Euclidean distance, and $\alpha$ is the predefined margin. The LiDAR branch is trained following the same procedure. This stage ensures that each branch acquires a robust intra-modal retrieval capability, providing a well-conditioned initialization for the subsequent discovery of cross-modal commonalities.

\textbf{The Task-Specific Asymmetry Observation:} To inform the choice of alignment paradigm, we adopt conditional entropy as a diagnostic heuristic for analyzing the information-sharing pattern of the features. We treat the LiDAR/radar feature maps $\mathcal{F}^{\text{li}}$ and $\mathcal{F}^{\text{ra}}$ as random variables $L$ and $R$. Discretizing each feature space into $B$ bins, we estimate the joint distribution $P(L, R)$ via a normalized 2D histogram, and define the two conditional entropies as:
\begin{align}
H(L|R) = -\sum_{l \in \mathcal{X}_{\text{L}}} \sum_{r \in \mathcal{X}_{\text{R}}} P(l, r) \log \frac{P(l, r)}{P(r)} \\
H(R|L) = -\sum_{l \in \mathcal{X}_{\text{L}}} \sum_{r \in \mathcal{X}_{\text{R}}} P(l, r) \log \frac{P(l, r)}{P(l)}
\end{align}
where $l \in \mathcal{X}_{\text{L}}$ and $r \in \mathcal{X}_{\text{R}}$ are the discretized realizations, $P(l, r)$ is the joint probability, and $P(l)$, $P(r)$ are the marginal probabilities. Furthermore, given the identity $H(L|R)-H(R|L) = H(L)-H(R)$, the conditional entropy gap directly characterizes their marginal entropy difference. This allows us to link the directional property to the inherent characteristics of multimodal features.

\begin{figure}[t]
%\vspace{0.2cm}
  \centering
  \includegraphics[width=1\linewidth]{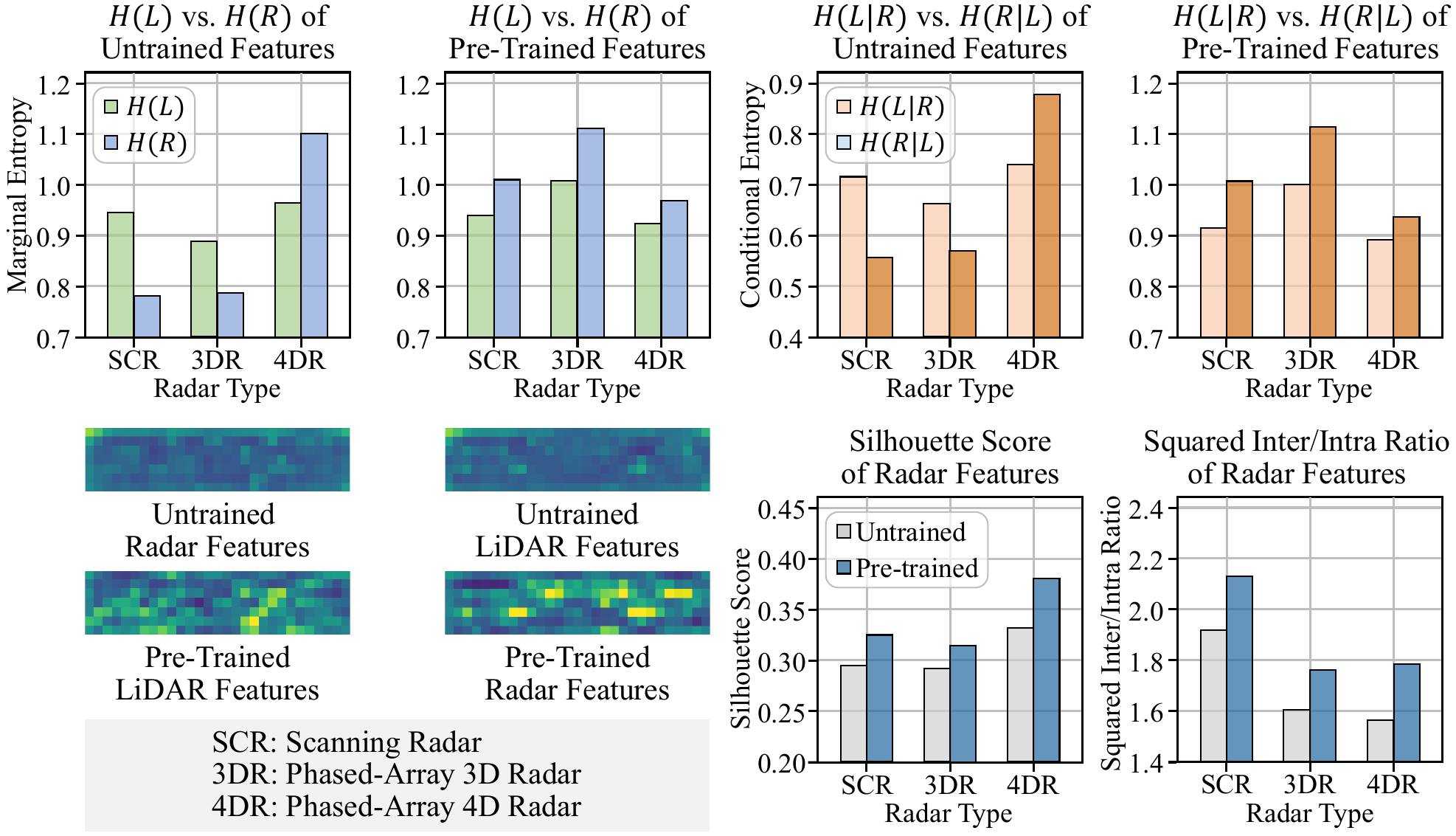}
  \caption{Representational evolution under pre-training. Modality-specific place recognition pre-training induces relatively larger radar entropy gain with concurrent gains in separability. The polar BEV feature visualizations are also provided.}
  \label{fig:entropy}
  \vspace{-0.65cm}
\end{figure}

We categorize the evaluation splits by radar type: scanning radar, phased-array 3D radar, and phased-array 4D radar. $H(L|R)$, $H(R|L)$, $H(L)$ and $H(R)$ are averaged over the corresponding splits. As shown in~\figref{fig:entropy}, place recognition pre‑training consistently leads to the trend $H(L|R)<H(R|L)$, driven by a relatively larger increase in $H(R)$. The phased-array 4D variant already satisfies $H(L|R)<H(R|L)$ at initialization and also preserves the ordering. The trend extends beyond means to individual data splits, and remains qualitatively robust to the bin count $B$, the feature layer, or the random seeds. To further substantiate that the increase in $H(R)$ is not merely due to class-agnostic noise dispersion, we additionally probe the geometry of the learned radar descriptors using place-wise separability metrics, namely the squared inter/intra-place distance ratio and the silhouette score~\cite{rousseeuw1987silhouettes}. As shown in~\figref{fig:entropy}, the comparatively greater increase in $H(R)$ is accompanied by an increase in separability, suggesting that radar may benefit from maintaining a higher-entropy manifold whose variability is organized into discriminative structures, whereas LiDAR tends to form a relatively compact representation. This observation provides a heuristic for our choice of alignment direction.

\textbf{Asymmetric Alignment:} Motivated by the task-specific asymmetry observations, we perform asymmetric alignment on the pre-trained features. Specifically, we designate the radar branch as a frozen feature anchor and the LiDAR branch as a trainable student, and utilize the asymmetric InfoNCE loss~\cite{oord2019infonce} to align both local and global descriptors. This process is formulated as:
\begin{align}
\mathcal{L}_{\text{INl}}\!=\!-\frac{1}{N} \sum_{a=1}^{N} \log \frac{\exp(\{\mathcal{D}_{\text{loc}}^{\text{ra}}\}_{a}\!\cdot\!\{\mathcal{D}_{\text{loc}}^{\text{li}}\}_{a}/\tau)}{\sum_{b=1}^{N} \exp(\{\mathcal{D}_{\text{loc}}^{\text{ra}}\}_{a}\!\cdot\!\{\mathcal{D}_{\text{loc}}^{\text{li}}\}_{b} / \tau)} \\
\mathcal{L}_{\text{INg}}\!=\!-\frac{1}{N} \sum_{a=1}^{N} \log \frac{\exp(\{\mathcal{D}_{\text{glob}}^{\text{ra}}\}_{a}\!\cdot\!\{\mathcal{D}_{\text{glob}}^{\text{li}}\}_{a} / \tau)}{\sum_{b=1}^{N} \exp(\{\mathcal{D}_{\text{glob}}^{\text{ra}}\}_{a}\!\cdot\!\{\mathcal{D}_{\text{glob}}^{\text{li}}\}_{b} / \tau)}
\end{align}
where $\{\mathcal{D}_{\text{loc}}^{\text{ra}}\}$, $\{\mathcal{D}_{\text{loc}}^{\text{li}}\}$, $\{\mathcal{D}_{\text{glob}}^{\text{ra}}\}$, $\{\mathcal{D}_{\text{glob}}^{\text{li}}\}$ denote batches of descriptors, $N$ represents the batch size, and $\tau$ is a predefined temperature parameter. %By leveraging InfoNCE, we maximize  mutual information between modalities. Under our asymmetric strategy, since $H(R)$ is fixed, this is equivalent to minimizing the conditional entropy $H(R|L)$. This targeted optimization effectively resolves the primary uncertainty bottleneck, exploiting shared geometry while preserving feature discriminability. 
Furthermore, we observe that asymmetric alignment generally outperforms joint fine-tuning of both branches (see~\tabref{tab:alignment}). This superiority can be attributed to the establishment of a discriminative anchor, which prevents the modalities from over-aligning into a shared latent space and thereby sacrificing task-specific saliency (see~\tabref{tab:alignment_uni}).

%%%%%%%%%%%%%%%%%%%%%%%%%%%%%%%%%%%%%%%%%%%%%%%%%%%%%%%%%%%%%%%%%%%%%%%%%%%%%%%%
\section{Experiments}
\label{sec:experiments}

\subsection{Experimental Setup}
\label{sec:exp_setup}

\textbf{Datasets:} We evaluate the proposed framework across five public datasets, namely MulRan~\cite{kim2020mulran}, Boreas~\cite{burnett2023boreas}, nuScenes~\cite{caesar2020nuscenes}, Snail-Radar~\cite{huai2025snailradar} and HeRCULES~\cite{kim2025hercules}. For scanning radar-to-LiDAR place recognition, we employ MulRan as the primary benchmark. We follow the data split from~\cite{qi2025lrfusionpr} but additionally include `KAIST 01' and `KAIST 02' in the training set. Furthermore, we conduct zero-shot evaluation on Boreas to assess cross-dataset generalization. To ensure a strict distribution shift, all Boreas data are strictly excluded from training to ensure the model generalizes to unseen sensors and locations. We define two distinct scenarios, Bor-Clear and Bor-Snowy. Both benchmarks utilize the ‘2020-12-18-13-44’ sequence as the database, while the query sequences are ‘2021-09-14-20-00’ and ‘2021-01-26-11-22’, respectively. For phased-array 3D radar-to-LiDAR evaluation, we adopt nuScenes with the data splits defined in~\cite{cait2022autoplace, qi2025lrfusionpr}, where the evaluated 3D radar is Continental ARS-408. We benchmark phased-array 4D radar-to-LiDAR performance using Snail-Radar, following TransLoc4D~\cite{peng2024transloc4d} while excluding non-autonomous-driving scenarios. The HeRCULES dataset is also utilized for 4D radar-to-LiDAR evaluation, with the `RiverIsland 01-02' sequences used for training and `SportsComplex 01-02' for testing.

\textbf{Implementation Details:} We set the size of both radar and LiDAR polar BEV to $(225, 50)$.  
For 4D radar, we generate polar BEVs of size $(100, 100)$. The maximum perception range $m$ is fixed at 80 meters. This relatively low polar BEV resolution ensures real-time performance while also providing an inherent sub-grid tolerance to small extrinsic calibration errors. Both $\mathcal{D}_{\text{loc}}$ and $\mathcal{D}_{\text{glob}}$ have a dimension of $D=256$, resulting in the final descriptor of size $1 \times 512$. The bin count is set to $B = 10$. For pre-training, we set $\alpha = 0.5$ and $J = 10$, with positives and negatives defined by distances within 9 meters and beyond 12 meters, respectively. The asymmetric alignment stage uses $N=12$ and $\tau = 0.07$. Both stages use the Adam optimizer with an initial learning rate of $5 \times 10^{-5}$ and 20\% epoch-wise decay. Following~\cite{cait2022autoplace}, the retrieval is considered successful if the candidate falls within 9 meters of the query. For the HeRCULES dataset experiments, we initialize RLPR with the pre-trained weights from Snail-Radar and perform fine-tuning for one epoch.

\textbf{Baselines:} We compare RLPR against three categories of baselines. The first comprises dedicated R2L methods, Radar-to-LiDAR~\cite{yin2021radar} and RaLF~\cite{nayak2024ralf}, which are designed for scanning radar and thus evaluated only in corresponding datasets. The second involves SOTA unimodal methods adapted to the cross-modal setting, denoted with a ``-CM'' suffix. Their original architectures are preserved but are retrained on cross-modal pairs using their respective losses. Specifically, KidnappedRadar~\cite{suaftescu2020kidnapped}, AutoPlace~\cite{cait2022autoplace} and TransLoc4D~\cite{peng2024transloc4d} are utilized as baselines only for their target radar types (scanning, phased-array 3D and 4D radar, respectively), while BEVPlace++~\cite{luo2024bevplace++} is evaluated across all datasets. The third category comprises C2L methods LIP-Loc~\cite{puligilla2024liploc} and I2P-CMPR~\cite{yao2025i2pcmpr}, whose range image-to-RGB alignment paradigm generalizes naturally to the polar BEV-based R2L task. For LIP-Loc, inputs are replaced with polar BEVs while preserving the original training pipeline. For I2P-CMPR, the Multi-View Generation module designed for cross-modal field-of-view misalignment is removed from the architecture, and the corresponding Multi-View Loss is omitted during training. Both methods are evaluated across all datasets. All baselines are implemented using their open-source code and retrained on our splits to ensure a rigorous comparison. For RaLF, we utilize the official pre-trained weights, which were jointly optimized across multiple datasets. This evaluation protocol is designed to demonstrate that our method, even with more restricted training data, can achieve superior cross-dataset generalization compared to the fully-optimized SOTA R2L baseline.

\begin{table*}[ht]
  \centering
  	\vspace{0.0cm}
  	\setlength{\tabcolsep}{4.8pt}
  	\renewcommand\arraystretch{0.95}
    \setlength{\abovecaptionskip}{0.15cm}
    \caption{Scanning radar-to-LiDAR performance on MulRan \& Boreas}
    \resizebox{\textwidth}{!}{
    \begin{tabular}{>{\centering\arraybackslash}m{3cm}|cccc|cccc|cccc}
      \toprule
      \multirow{2}{*}{Methods} &\multicolumn{4}{c|}{MulRan-Riverside}&\multicolumn{4}{c|}{MulRan-DCC}&\multicolumn{4}{c}{Bor-Clear} \\ \cline{2-13}
      ~ & AR@1 & AR@5 & AR@10 & max $F_{1}$ & AR@1 & AR@5 & AR@10  & max $F_{1}$ & AR@1 & AR@5 & AR@10  & max $F_{1}$ \\ \hline
      KidnappedRadar-CM~\cite{suaftescu2020kidnapped} & 5.452 & 12.94 & 20.30 & 0.1286 & 9.575 & 23.94 & 42.02 & 0.2222 & 0.6890 & 2.493 & 4.133 & 0.02755 \\
      BEVPlace++-CM~\cite{luo2024bevplace++} & 2.668 & 10.21 & 16.94 & 0.05198 & 2.660 & 10.64 & 21.81 & 0.07207 & 0.6562 & 2.067 & 3.642 & 0.01377 \\
      LIP-Loc~\cite{puligilla2024liploc} & 7.135 & 17.40 & 23.43 & 0.1024 & 9.574 & 26.06 & 39.36 & 0.4000 & 0.7532 & 2.636 & 4.759 & 0.005448 \\
      I2P-CMPR~\cite{yao2025i2pcmpr} & 4.292 & 19.84 & 31.73 & 0.01497 & 5.851 & 21.81 & 31.38 & 0.08163 & 3.771 & 12.12 & 19.58 & 0.01304\\
      Radar-to-LiDAR~\cite{yin2021radar} & \underline{31.13} & \underline{44.54} & \underline{50.96} & 0.4748 & 12.77 & 24.47 & 32.98 & 0.2264 & 24.51 & 49.61 & 61.25 & 0.3938 \\
      RaLF~\cite{nayak2024ralf} & 30.10 & 38.92 & 45.88 & \underline{0.5942} &
      \underline{34.04} & \underline{50.53} & \underline{55.85} & \underline{0.5224} & \underline{73.85} & \underline{81.56} & \underline{85.37} & \underline{0.8576}\\
      \rowcolor{gray!25}
      RLPR (ours) & \textbf{64.85} & \textbf{78.77} & \textbf{84.22} & \textbf{0.8057} &
      \textbf{55.32} & \textbf{73.94} & \textbf{82.98} & \textbf{0.7419} & \textbf{87.60} & \textbf{94.26} & \textbf{95.31} & \textbf{0.9469} \\
      \bottomrule
    \multicolumn{13}{p{0.9\linewidth}}{The best and secondary results are highlighted in \textbf{bold black} and \underline{underline} respectively.}\\
    \end{tabular}
    }
    \label{tab:scanning}
    \vspace{-0.6cm}
\end{table*}

\begin{table*}[ht]
  \centering
  	\vspace{0.3cm}
  	\setlength{\tabcolsep}{4.8pt}
  	\renewcommand\arraystretch{0.95}
    \setlength{\abovecaptionskip}{0.15cm}
    \caption{Phased-array 3D radar-to-LiDAR performance on nuScenes}
    \resizebox{\textwidth}{!}{
    \begin{tabular}{>{\centering\arraybackslash}m{3cm}|cccc|cccc|cccc}
      \toprule
      \multirow{2}{*}{Methods} &\multicolumn{4}{c|}{BS}&\multicolumn{4}{c|}{SON}&\multicolumn{4}{c}{SQ} \\ \cline{2-13}
      ~ & AR@1 & AR@5 & AR@10 & max $F_{1}$ & AR@1 & AR@5 & AR@10  & max $F_{1}$ & AR@1 & AR@5 & AR@10  & max $F_{1}$ \\ \hline
      AutoPlace-CM~\cite{cait2022autoplace} & \underline{64.65} & \underline{76.48} & \underline{79.51} & \underline{0.7890} & 0.9072 & 1.693 & 2.510 & 0.01822 & 0.4706 & 2.353 & 3.177 & 0.03540 \\
      BEVPlace++-CM~\cite{luo2024bevplace++} & 1.380 & 5.087 & 8.333 & 0.07576 & 0.7557 & 3.023 & 5.441 & 0.04545 & 4.935 & 16.80 & 25.73 & 0.09524 \\
      LIP-Loc~\cite{puligilla2024liploc} & 14.64 & 30.07 & 37.94 & 0.2615 & 2.328 & 8.620 & 14.18 & 0.07122 & 5.882 & 18.45 & 26.79 & 0.1681 \\
      I2P-CMPR~\cite{yao2025i2pcmpr} & 12.31 & 26.17 & 34.45 & 0.2635 & \underline{15.84} & \underline{34.07} & \underline{42.65} & \underline{0.3192} & \underline{18.45} & \underline{38.43} & \underline{48.06} & \underline{0.3518}\\
      \rowcolor{gray!25}
      RLPR (ours) & \textbf{70.85} & \textbf{80.30} & \textbf{82.79} & \textbf{0.8907} & \textbf{59.31} & \textbf{77.42} & \textbf{83.07} & \textbf{0.7582} & \textbf{47.71} & \textbf{60.63} & \textbf{66.04} & \textbf{0.6893} \\
      \bottomrule
    \end{tabular}
    }
    \label{tab:singlechip}
    \vspace{-0.65cm}
\end{table*}

% \begin{table}[t]
%   \centering
%   \vspace{0.15cm}
%   \setlength{\tabcolsep}{4pt}
%   \renewcommand\arraystretch{0.95}
%   \setlength{\abovecaptionskip}{0.15cm}
%   \caption{4D radar-to-LiDAR performance on Snail-Radar}
%   \resizebox{\columnwidth}{!}{
%   \footnotesize{
%     \begin{tabular}{c|ccccc}
%         \toprule
%         \multirow{2}{*}{Method} & \multicolumn{5}{c}{81r} \\ \cline{2-6}
%         ~ & AR@1 & AR@5 & AR@10 & AR@20 & max $F_{1}$ \\ \hline
%         TransLoc4D-CM~\cite{peng2024transloc4d} & 13.73 & \underline{19.82} & 23.18 & 26.84 & 0.4015 \\
%         BEVPlace++-CM~\cite{luo2024bevplace++} & 0.4884 & 0.5215 & 0.5463 & 1.010 & 0.04369 \\
%         LIP-Loc~\cite{puligilla2024liploc} & 7.427 & 18.00 & \underline{25.68} & \underline{35.25} & 0.1319 \\
%         I2P-CMPR~\cite{yao2025i2pcmpr} & \underline{16.26} & 17.51 & 18.57 & 24.58 & \underline{0.4118} \\
%         \rowcolor{gray!25}
%         RLPR (ours) & \textbf{44.71} & \textbf{47.58} & \textbf{50.16} & \textbf{62.13} & \textbf{0.6304} \\
%         \bottomrule
%     \end{tabular}
%     }
%   }
%   \label{tab:4dradar}
%   \vspace{-0.6cm}
% \end{table}

\begin{table}[t]
  \centering
  \vspace{0.05cm}
  \setlength{\tabcolsep}{2pt}
  \renewcommand\arraystretch{0.95}
  \setlength{\abovecaptionskip}{0.15cm}
  \caption{Phased-array 4D radar-to-LiDAR performance}
  \resizebox{\columnwidth}{!}{
  \footnotesize{
    \begin{tabular}{c|ccc|ccc}
        \toprule
        \multirow{2}{*}{Method} & \multicolumn{3}{c|}{81r} & \multicolumn{3}{c}{SportsComplex} \\ \cline{2-7}
        ~ & AR@1 & AR@5 & max $F_{1}$ & AR@1 & AR@5 & max $F_{1}$ \\ \hline
        TransLoc4D-CM~\cite{peng2024transloc4d} & 13.73 & \underline{19.82} & 0.4015 & 21.35 & 33.43 & 0.5437 \\
        BEVPlace++-CM~\cite{luo2024bevplace++} & 0.4884 & 0.5215 & 0.04369 & 2.879 & 6.930 & 0.09195\\
        LIP-Loc~\cite{puligilla2024liploc} & 7.427 & 18.00 & 0.1319 & 10.38 & 26.07 & 0.1796\\
        I2P-CMPR~\cite{yao2025i2pcmpr} & \underline{16.26} & 17.51 & \underline{0.4118} & \underline{32.40} & \underline{39.88} & \underline{0.6049} \\
        \rowcolor{gray!25}
        RLPR (ours) & \textbf{44.71} & \textbf{47.58} & \textbf{0.6304} & \textbf{72.16} & \textbf{82.62} & \textbf{0.8418}\\
        \bottomrule
    \end{tabular}
    }
  }
  \label{tab:4dradar}
  \vspace{-0.6cm}
\end{table}

\subsection{Radar-to-LiDAR Place Recognition Evaluation}
\label{sec:r2l_eval}
Following~\cite{cait2022autoplace, qi2025lrfusionpr}, we report the average recall of the top-K candidates (AR@K) and max $F_{1}$ score as primary metrics to evaluate cross-modal recognition accuracy. We present the R2L recognition accuracy comparisons for scanning radar in~\tabref{tab:scanning}, and for phased-array 3D and 4D radars in~\tabref{tab:singlechip} and~\tabref{tab:4dradar}, respectively. Note that the sequences `Riverside', `DCC', `SON', `SQ', and `SportsComplex' represent intra-dataset generalization scenarios with no spatiotemporal overlap with the training set, while Bor-Clear serves as the cross-dataset generalization scenario. Our proposed RLPR consistently outperforms the baselines, underscoring its robustness across heterogeneous radar types. Notably, on the cross-dataset Bor-Clear sequence, RLPR exhibits superior zero-shot generalization capabilities compared to RaLF, which is trained on more data. Furthermore, the performance on the SportsComplex split validates that RLPR adapts to heterogeneous 4D radars with simple fine-tuning, underscoring the generalization capability of the learned features.

\subsection{Unimodal Place Recognition Evaluation}
\label{sec:unim_eval}
\begin{figure}[t]
\vspace{0.1cm}
  \centering
  \includegraphics[width=1\linewidth]{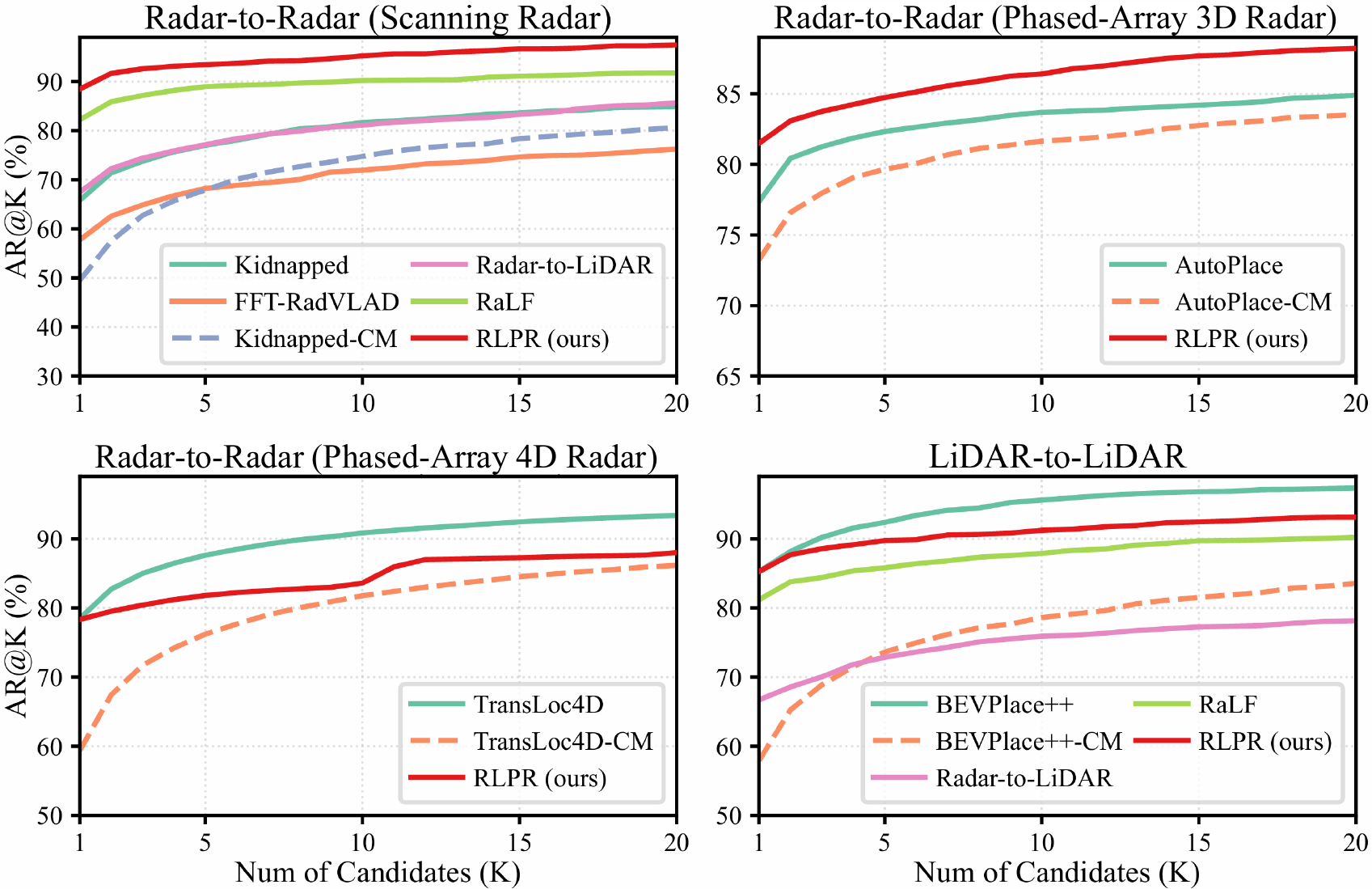}
  \caption{A comparative visualization of radar-to-radar and LiDAR-to-LiDAR place recognition accuracy.}
  \label{fig:unimodal}
  \vspace{-0.75cm}
\end{figure}

In this section, we evaluate the RPR and LPR performance, utilizing only the corresponding unimodal branch of the trained network for evaluation. We report the average AR@K across all evaluation sequences for each radar type, alongside the mean LPR performance on MulRan and Boreas. As illustrated in~\figref{fig:unimodal}, our RPR performance not only surpasses R2L baselines, but also achieves comparable accuracy with RPR methods. Furthermore, while the LiDAR branch exhibits a slight performance gap compared to SOTA LPR method, it outperforms all R2L baselines. These results demonstrate that, benefiting from the proposed TACMA strategy, our network achieves robust cross-modal alignment without compromising unimodal discriminability.

\begin{table*}[ht]
  \centering
  	\vspace{0.0cm}
  	\setlength{\tabcolsep}{2.7pt}
  	\renewcommand\arraystretch{0.95}
    \setlength{\abovecaptionskip}{0.15cm}
    \caption{Ablation study on place recognition pre-training}
    \resizebox{\textwidth}{!}{
    \begin{tabular}{>{\centering\arraybackslash}m{3.7cm}|cccccc|cccccc|cc}
      \toprule
      \multirow{2}{*}{\makecell{Strategy}} &\multicolumn{2}{c}{BS}&\multicolumn{2}{c}{SON}&\multicolumn{2}{c|}{SQ}&\multicolumn{2}{c}{Riverside}&\multicolumn{2}{c}{DCC}&\multicolumn{2}{c|}{Bor-Clear}&\multicolumn{2}{c}{81r} \\ \cline{2-15}
      ~ & AR@1 & AR@5 & AR@1 & AR@5 & AR@1 & AR@5 & AR@1 & AR@5 & AR@1 & AR@5 & AR@1 & AR@5 & AR@1 & AR@5 \\ \hline
      No Pre-training & 20.35 & 39.69 & 5.804 & 15.39 & 3.760 & 12.34 & 8.469 & 19.55 & 10.64 & 35.11 & 3.018 & 5.053 & 29.85 & 32.21 \\
      Joint Optimization & 44.18 & 60.90 & 24.88 & 38.15 & 20.21 & 33.37 & 21.64 & 39.56 & 22.87 & 48.40 & 17.03 & 21.95 & 32.83 & 34.29 \\
      Anchor-Only Pre-Training & 60.14 & 75.07 & 22.43 & 42.50 & 23.74 & 42.19 & 34.17 & 55.51 & 32.98 & 60.64 & 30.87 & 39.07 & 28.04 & 30.60 \\
      \rowcolor{gray!25}
      Modality-Specific Pre-Training & \textbf{70.85} & \textbf{80.30} & \textbf{59.31} & \textbf{77.42} & \textbf{47.71} & \textbf{60.63} & \textbf{64.85} & \textbf{78.77} & \textbf{55.32} & \textbf{73.94} & \textbf{87.60} & \textbf{94.26} & \textbf{44.71} & \textbf{47.58} \\
      \bottomrule
    \end{tabular}
    }
    \label{tab:pretraining}
    \vspace{-0.65cm}
\end{table*}

\begin{table*}[ht]
  \centering
  	\vspace{0.35cm}
  	\setlength{\tabcolsep}{2.7pt}
  	\renewcommand\arraystretch{0.95}
    \setlength{\abovecaptionskip}{0.15cm}
    \caption{Ablation study on alignment strategy}
    \resizebox{\textwidth}{!}{
    \begin{tabular}{>{\centering\arraybackslash}>{\centering\arraybackslash}m{1.2cm}>{\centering\arraybackslash}m{2.3cm}|cccccc|cccccc|cc}
      \toprule
      \multirow{2}{*}{Loss} &\multirow{2}{*}{\makecell{Alignment \\ Strategy}} &\multicolumn{2}{c}{BS}&\multicolumn{2}{c}{SON}&\multicolumn{2}{c|}{SQ}&\multicolumn{2}{c}{Riverside}&\multicolumn{2}{c}{DCC}&\multicolumn{2}{c|}{Bor-Clear}&\multicolumn{2}{c}{81r} \\ \cline{3-16}
      ~ & ~ & AR@1 & AR@5 & AR@1 & AR@5 & AR@1 & AR@5 & AR@1 & AR@5 & AR@1 & AR@5 & AR@1 & AR@5 & AR@1 & AR@5 \\ \hline
      Triplet & Frozen L & 44.29 & 58.09 & 32.22 & 49.58 & 7.451 & 23.40 & 46.40 & 64.10 & 45.74 & 69.68 & 62.60 & 68.41 & 17.95 & 19.49 \\
      Triplet & Both Trainable & 61.09 & 71.37 & 44.53 & 61.22 & 45.71 & 58.05 & 56.38 & 72.85 & 56.38 & \textbf{78.72} & 84.84 & 88.16 & 26.05 & 27.94 \\
      Triplet & Frozen R & 55.89 & 66.90 & 37.88 & 58.52 & 44.18 & 56.64 & 59.05 & 75.00 & 55.32 & 75.00 & 84.71 & 88.19 & 23.45 & 25.05 \\ \hline
      MSE & Frozen L & 60.08 & 71.18 & 27.81 & 42.50 & 39.84 & 56.05 & 56.61 & 71.29 & 39.89 & 66.49 & 46.10 & 52.76 & 30.67 & 32.60 \\
      MSE & Both Trainable & 1.735 & 6.478 & 1.183 & 4.414 & 2.473 & 7.173 & 11.48 & 23.67 & 10.11 & 24.47 & 0.7218 & 2.067 & 1.349 & 1.606 \\
      MSE & Frozen R & 68.43 & 77.30 & 57.89 & 74.24 & 46.77 & 59.11 & 62.88 & 77.90 & \textbf{56.91} & 75.53 & 80.94 & 89.90 & 43.10 & 45.94 \\ \hline
      InfoNCE & Frozen L & 65.60 & 75.75 & 38.60 & 54.75 & 43.60 & 56.17 & 64.68 & \textbf{80.34} & 51.60 & 77.66 & 69.49 & 74.54 & 40.64 & 42.74 \\
      InfoNCE & Both Trainable & 52.42 & 66.20 & 54.63 & 73.73 & 42.07 & 56.64 & 48.67 & 66.13 & 50.53 & 71.81 & 70.51 & 76.90 & 41.45 & \textbf{49.66} \\
      \rowcolor{gray!25}
      InfoNCE & Frozen R & \textbf{70.85} & \textbf{80.30} & \textbf{59.31} & \textbf{77.42} & \textbf{47.71} & \textbf{60.63} & \textbf{64.85} & 78.77 & 55.32 & 73.94 & \textbf{87.60} & \textbf{94.26} & \textbf{44.71} & 47.58 \\
      \bottomrule
    \end{tabular}
    }
    \label{tab:alignment}
    \vspace{-0.65cm}
\end{table*}

\begin{table*}[ht]
  \centering
  	\vspace{0.35cm}
  	\setlength{\tabcolsep}{2.0pt}
  	\renewcommand\arraystretch{0.95}
    \setlength{\abovecaptionskip}{0.15cm}
    \caption{The impact of alignment strategies on modality-specific discriminability}
    \resizebox{\textwidth}{!}{
    \begin{tabular}{c|cccc|cccc|cccc}
      \toprule
      \multirow{2}{*}{\makecell{Alignment \\ Strategy}} &\multicolumn{4}{c|}{Phased-Array 3D Radar (On Average)}&\multicolumn{4}{c|}{Scanning Radar (On Average)}&\multicolumn{4}{c}{Phased-Array 4D Radar (On Average)} \\ \cline{2-13}
      ~ & $H(L)$ & $H(R)$ & LPR (AR@1) & RPR (AR@1) & $H(L)$ & $H(R)$ & LPR (AR@1) & RPR (AR@1) & $H(L)$ & $H(R)$ & LPR (AR@1) & RPR (AR@1)  \\ \hline
      Pre-Alignment & 0.9378 & 1.088 & 88.26 & 81.51 & 1.060 & 1.184 & 86.78 & 88.14 & 0.9213 & 0.9670 & 87.34 & 78.45 \\
      Frozen L & 0.9378 & 1.114 & 88.26 (-0.00) & 81.50 (\textcolor{red}{-0.01}) & 1.060 & 1.182 & 86.78 (-0.00) & 85.81 (\textcolor{red}{-2.33}) & 0.9213 & 1.055 & 87.34 (-0.00) & 73.10 (\textcolor{red}{-5.35}) \\
      Both Trainable & 1.215 & 1.112 & 81.82 (\textcolor{red}{-6.44}) & 74.00 (\textcolor{red}{-7.51}) & 1.110 & 1.103 & 79.58 (\textcolor{red}{-7.20}) & 77.39 (\textcolor{red}{-10.75}) & 1.097 & 0.9365 & 82.72 (\textcolor{red}{-4.62}) & 76.91 (\textcolor{red}{-1.54}) \\
      \rowcolor{gray!25}
      Frozen R & 1.266 & 1.088 & 85.41 (\textcolor{red}{-2.85}) & 81.51 (-0.00) & 1.156 & 1.184 & 85.23 (\textcolor{red}{-1.55}) & 88.14 (-0.00) & 1.127 & 0.9670 & 83.70 (\textcolor{red}{-3.64}) & 78.45 (-0.00) \\
      \bottomrule
    \end{tabular}
    }
    \label{tab:alignment_uni}
    \vspace{-0.65cm}
\end{table*}

\begin{table}[t]
  \centering
  \vspace{0.1cm}
  \setlength{\tabcolsep}{2pt}
  \renewcommand\arraystretch{0.95}
  \setlength{\abovecaptionskip}{0.15cm}
  \caption{Ablation study on network architecture}
  \resizebox{\columnwidth}{!}{
  \footnotesize{
    \begin{tabular}{ccccccc|cccc}
        \toprule
        \multirow{2}{*}{CMP} & \multirow{2}{*}{GAP} & \multirow{2}{*}{CAP} & \multirow{2}{*}{LD} & \multirow{2}{*}{GD} & \multirow{2}{*}{PCE} & \multirow{2}{*}{PCG} & \multicolumn{2}{c}{BS}  & \multicolumn{2}{c}{SON} \\ \cline{8-11}
        ~ & ~ & ~ & ~ & ~ & ~ & ~ & AR@1 & AR@5 & AR@1 & AR@5 \\ \hline
        \ding{55} & \ding{55} & \ding{51} & \ding{51} & \ding{55} & \ding{55} & \ding{55} & 43.21 & 61.01 & 8.827 & 19.56 \\
        \ding{55} & \ding{55} & \ding{55} & \ding{55} & \ding{51} & \ding{55} & \ding{55} & 51.00 & 69.07 & 9.341 & 23.49 \\
        \ding{55} & \ding{55} & \ding{51} & \ding{51} & \ding{51} & \ding{55} & \ding{55} & 64.26 & 75.70 & 55.17 & 71.70 \\
        \ding{51} & \ding{55} & \ding{55} & \ding{51} & \ding{51} & \ding{51} & \ding{55} & 68.21 & 78.33 & 49.03 & 67.71 \\
        \ding{55} & \ding{51} & \ding{55} & \ding{51} & \ding{51} & \ding{51} & \ding{55} & 58.28 & 72.46 & 16.44 & 30.50 \\
        \ding{55} & \ding{55} & \ding{51} & \ding{51} & \ding{51} & \ding{55} & \ding{51} & 65.61 & 76.43 & 55.67 & 73.78 \\
        \rowcolor{gray!25}
        \ding{55} & \ding{55} & \ding{51} & \ding{51} & \ding{51} & \ding{51} & \ding{55} & \textbf{70.85} & \textbf{80.30} & \textbf{59.31} & \textbf{77.42} \\
        \bottomrule
    \end{tabular}
    }
    }
  \label{tab:architecture}
  \vspace{-0.4cm}
\end{table}

\subsection{Ablation Studies}
\label{sec:ablation}

\textbf{Place Recognition Pre-Training:} We conduct an ablation study on the pre-training stage, comparing our proposed strategy against three variants: No Pre-training, which bypasses the pre-training stage; Joint Optimization, in which unimodal and cross-modal losses are optimized simultaneously; and Anchor-Only Pre-Training, which optimizes only the teacher branch following~\cite{cattaneo2020global}. As shown in~\tabref{tab:pretraining}, our Modality-Specific Pre-Training achieves substantial accuracy gains over other variants, indicating that constructing discriminative feature spaces for both modalities prior to cross-modal alignment is essential.

\begin{table}[t]
  \centering
  \vspace{0.1cm}
  \setlength{\tabcolsep}{4pt}
  \renewcommand\arraystretch{0.95}
  \setlength{\abovecaptionskip}{0.15cm}
  \caption{Comparison of performance in snowy scenario}
  \resizebox{\columnwidth}{!}{
  \footnotesize{
    \begin{tabular}{c|ccccc}
        \toprule
        \multirow{2}{*}{Method} & \multicolumn{5}{c}{Bor-Snow} \\ \cline{2-6}
        ~ & AR@1 & AR@5 & AR@10 & AR@20 & max $F_{1}$ \\ \hline
        MinkLoc3Dv2~\cite{komorowski2022improving} & 18.83 & 31.63 & 37.45 & 43.49 & 0.4606 \\
        BEVPlace++~\cite{luo2024bevplace++} & 67.31 & 73.62 & 76.01 & 79.49 & 0.8524 \\
        Radar-to-LiDAR~\cite{yin2021radar} & 5.158 & 24.49 & 48.46 & 62.68 & 0.09880 \\
        RaLF~\cite{nayak2024ralf} & 74.79 & 86.81 & 90.65 & 93.88 & 0.8588 \\
        \rowcolor{gray!25}
        RLPR (ours) & \textbf{85.52} & \textbf{93.92} & \textbf{95.86} & \textbf{97.58} & \textbf{0.9226} \\
        \bottomrule
    \end{tabular}
    }
  }
  \label{tab:weather}
  \vspace{-0.7cm}
\end{table}

\textbf{Asymmetric Alignment:} To validate our alignment strategy, we conduct an ablation study across three regimes: frozen LiDAR branch (Frozen L), joint fine-tuning (Both Trainable), and frozen radar branch (Frozen R), benchmarked under Triplet, MSE, and InfoNCE losses. As shown in~\tabref{tab:alignment}, the combination of InfoNCE and Frozen R generally outperforms the other regimes, empirically corroborating our strategy. Notably, the advantage of Frozen R appears to scale with the alignment pressure. Under the Triplet loss, whose soft ranking constraint applies the weakest pressure, both branches can co-adapt freely, allowing Both Trainable to exceed Frozen R. Under the moderate pressure of InfoNCE, while Frozen L and Both Trainable retain marginal advantages on a few splits, yet both fall well below Frozen R on the cross-dataset Bor-Clear split where generalizable cross-modal patterns matter most. Under the rigid MSE constraint, joint optimization collapses, while Frozen R remains stable. These patterns indicate that the advantage of Frozen R becomes more pronounced both when rigorous cross-domain generalization is required and when the loss function imposes higher alignment pressure.

To further elucidate this behavior, \tabref{tab:alignment_uni} reports the joint evolution of marginal entropy and unimodal discriminability. Under Frozen L, $H(R)$ fails to decrease significantly, and rather than improving via ``denoising," RPR performance actually degrades. This further suggests that the increase in $H(R)$ may not be an artifact of sensor noise. In the Both Trainable regime, although conditional entropy evolves similarly to Frozen R, both LPR and RPR performance drop severely due to the lack of a discriminative anchor. Conversely, under Frozen R, the LiDAR branch accommodates the radar manifold with minimal loss in LPR accuracy while preserving RPR performance.

\begin{figure}[t]
\vspace{0.1cm}
  \centering
  \includegraphics[width=1\linewidth]{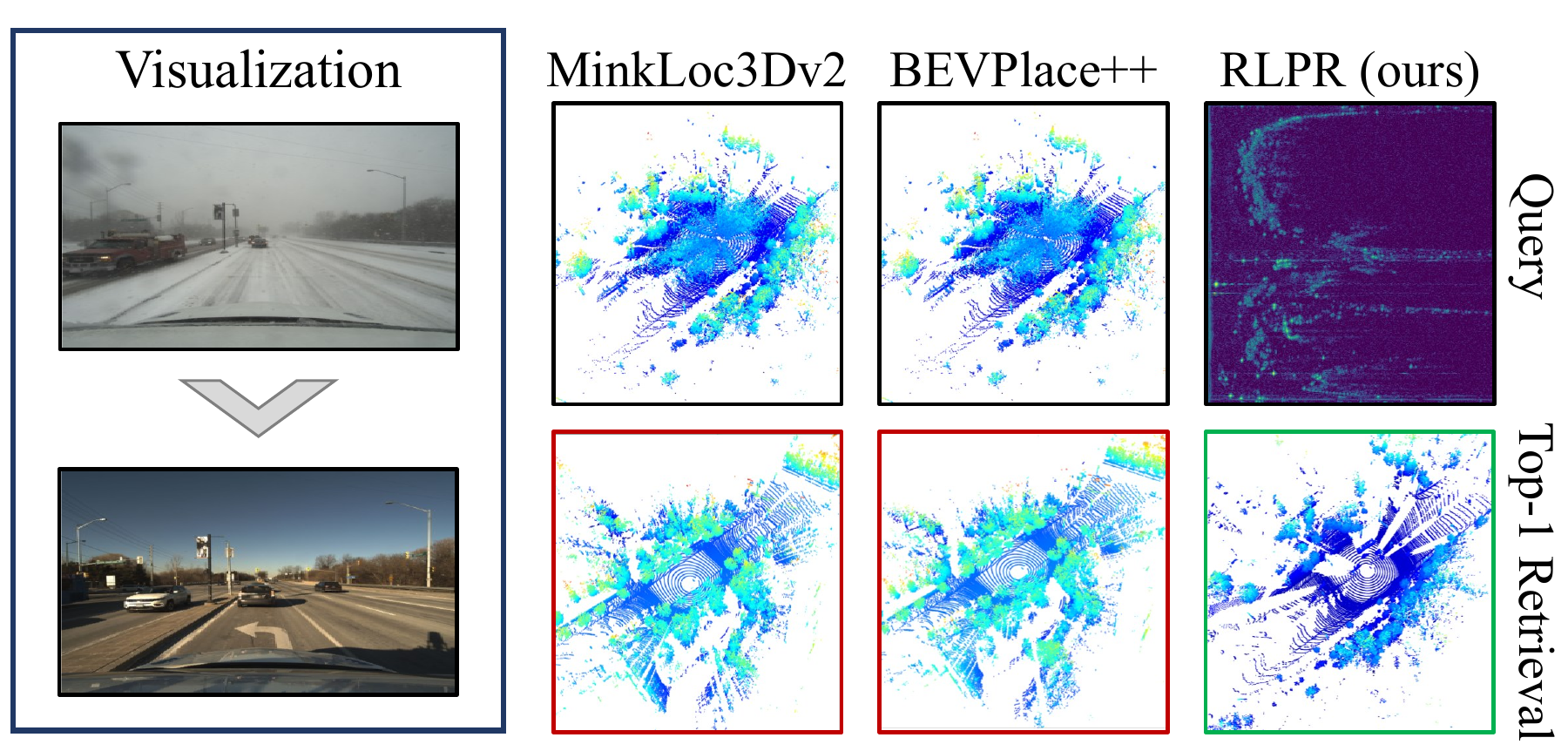}
  \caption{Visualization of retrieval results in adverse weather. LPR methods are severely degraded by snow-induced noise, leading to consistent retrieval failures, whereas our proposed R2L method remains robust.}
  \label{fig:viz_adverse}
  \vspace{-0.75cm}
\end{figure}

\textbf{Network Architecture:} We conduct an ablation study to evaluate the contribution of each core component. First, the Local Descriptor (LD) and Global Descriptor (GD) are ablated to assess their respective impacts. Furthermore, the Polar Context Enhancer (PCE) is evaluated by comparing our Mamba-based design against both its omission and a gating-based alternative (PCG). We also examine the choice of introducing channel-wise average pooling (CAP) for local descriptor extraction. Specifically, under strictly controlled conditions, we replace CAP with global average pooling (GAP) and channel-wise max pooling (CMP). As detailed in~\tabref{tab:architecture}, the LD and GD complement each other effectively, both being essential to ensure recognition accuracy. The incorporation of the PCE further enhances performance, and demonstrates the superiority of Mamba-based long-range modeling over the gating-based alternative. Finally, CAP significantly outperforms both CMP and GAP, justifying our architectural choices.

\subsection{Robustness to Adverse Weather}
\label{sec:weather}
In this section, we benchmark the robustness of RLPR on the Bor-Snow sequence, which is a zero-shot scenario for all methods. As shown in~\tabref{tab:weather}, SOTA LPR methods suffer severe performance degradation. This degradation stems from the snow-induced LiDAR artifacts, which lead to consistent retrieval failure patterns, as visualized in~\figref{fig:viz_adverse}. In contrast, our proposed R2L framework demonstrates superior resilience, indicating the potential of R2L as a complementary method for robust localization in harsh environments.

\subsection{Efficiency Analysis}
\label{sec:efficiency}
In this section, we evaluate the computational efficiency of our proposed method on the Bor-Snow split. Experiments are conducted on a system with an i7-14700KF CPU and an RTX 4060Ti GPU, with the results shown in~\tabref{tab:efficiency}. Notably, in an experimental mode that simulates real-world deployment by performing inference per query frame without parallelization, our method achieves an average descriptor extraction time of 2.88 ms and a top-1 retrieval time of 0.05 ms, demonstrating superior real-time performance. Beyond inference time, our proposed method maintains an acceptable parameter count while achieving significantly lower FLOPs and descriptor dimensions compared to R2L baselines, which illustrates its system-level efficiency.

\begin{table}[t]
  \centering
  \vspace{0.1cm}
  \setlength{\tabcolsep}{2.8pt}
  \renewcommand\arraystretch{0.95}
  \setlength{\abovecaptionskip}{0.15cm}
  \caption{System-level efficiency comparison}
  \resizebox{\columnwidth}{!}{
  \footnotesize{
    \begin{tabular}{c|cccccc}
        \toprule
        \multirow{2}{*}{Method} & Params & FLOPs & Desc. & Extract & Retrieval & Total \\
        ~ & (M) & (G) & Size & (ms) & (ms) & (ms) \\ \hline
        Radar-to-LiDAR~\cite{yin2021radar} & 4.3 & 1.11 & 4800 & 2.91 & 1.16 & 4.07 \\
        RaLF~\cite{nayak2024ralf} & \textbf{3.3} & 5.32 & 73728 & 3.25 & 21.36 & 24.61 \\
        \rowcolor{gray!25}
        RLPR (ours) & 4.5 & \textbf{0.40} & \textbf{512} & \textbf{2.88}  & \textbf{0.05} & 2.93  \\
        \bottomrule
    \end{tabular}
    }
  }
  \label{tab:efficiency}
  \vspace{-0.75cm}
\end{table}

\section{Conclusion}
\label{sec:conclusion}
In this paper, we propose RLPR, a R2L place recognition framework accommodating common radar types in autonomous driving. We introduce a tailored R2L network and the TACMA strategy for cross-modal alignment. Experiments across five datasets demonstrate its effectiveness as a robust localization solution for all-weather scenarios. Future research will explore its adaptation to low-cost single-chip radars, as well as unified representations that retain modality-specific cues such as Doppler velocity and RCS features.

% Only if applicable
%\section*{Acknowledgments}
%We thank XXX for fruitful discussions and for \dots

%% \clearpage
\bibliographystyle{ieeetran}

% All new citations should go to new.bib. The file glorified.bib should go
% be the one from the ipb server. After paper or related work has been
% written merge the entries from new.bib to glorified.bib ON THE SERVER,
% replace the glorified.bib in this repository and empty the new.bib
\footnotesize{
\bibliography{glorified, new}}

\end{document}